\pdfoutput=1

\documentclass[11pt]{article}

\usepackage[]{ACL}
\usepackage{times}
\usepackage{latexsym}

\usepackage[T1]{fontenc}

\usepackage[utf8]{inputenc}

\usepackage{microtype}

\usepackage{graphicx} 
\usepackage{booktabs}
\usepackage{amssymb}
\usepackage{amsmath}
\usepackage{multirow}
\usepackage[ruled,vlined]{algorithm2e}
\usepackage{comment}
\usepackage{listings}
\usepackage{pgfplots}
\usepackage{color}

\newcommand{\tabincell}[2]{\begin{tabular}{@{}#1@{}}#2\end{tabular}}

\definecolor{codegreen}{rgb}{0,0.6,0}
\definecolor{codegray}{rgb}{0.5,0.5,0.5}
\definecolor{codepurple}{rgb}{0.58,0,0.82}
\definecolor{backcolour}{rgb}{0.95,0.95,0.92}

\lstdefinelanguage{mypython}{morekeywords={def, return}}

\lstdefinestyle{mystyle}{
	language=mypython,
	commentstyle=\color{codegreen},
	keywordstyle=\color{codegreen},
	numberstyle=\color{codegray},
	stringstyle=\color{magenta},
	basicstyle=\ttfamily\fontsize{6.6pt}{7.92pt}\selectfont,
	breakatwhitespace=true,         
	breaklines=true,                 
	captionpos=b,                    
	keepspaces=true,                 
	numbers=none,                    
	showspaces=false,                
	showstringspaces=false,
	showtabs=false,                  
	tabsize=2,
	frame=single,
}

\title{EPiDA: An Easy Plug-in Data Augmentation Framework\\for High Performance Text Classification}

\author{
	Minyi Zhao$^{1}$ \quad Lu Zhang$^{1}$ \quad Yi Xu$^{1}$  \\{\bf Jiandong Ding}$^2$ \quad {\bf Jihong Guan}$^3$ \quad {\bf Shuigeng Zhou}$^1$\thanks{~~Corresponding author.} \\
	$^1$Shanghai Key Lab of Intelligent Information Processing,  and School of \\ Computer Science,  Fudan University, China\\
	$^2$Alibaba Group \quad $^3$Tongji University, China\\
	$^{1,2}$\{zhaomy20, l\_zhang19, yxu17, jdding, sgzhou\}@fudan.edu.cn \quad $^{3}$jhguan@tongji.edu.cn \\
}
\begin{document}
\maketitle
\begin{abstract}
Recent works have empirically shown the effectiveness of data augmentation~(DA) for NLP tasks, especially for those suffering from data scarcity.
Intuitively, given the size of generated data, their \emph{diversity} and \emph{quality} are crucial to the performance of targeted tasks. 
However, to the best of our knowledge, most existing methods consider only either the \emph{diversity} or the \emph{quality} of augmented data, thus cannot fully tap the potential of DA for NLP.   		
In this paper, we present an easy and plug-in data augmentation framework EPiDA to support effective text classification. EPiDA employs 
two mechanisms: \textit{relative entropy maximization} (REM) and \textit{conditional entropy minimization} (CEM) to control data generation, where REM is designed to enhance the diversity of augmented data while CEM is exploited to ensure their semantic consistency. EPiDA can support efficient and continuous data generation for effective classifier training. Extensive experiments show that EPiDA outperforms existing SOTA methods in most cases, though not using any agent network or pre-trained generation network, and it works well with various DA algorithms and classification models. Code is available at \url{https://github.com/zhaominyiz/EPiDA}.
\end{abstract}

\section{Introduction}
\noindent Data augmentation (DA) is widely-used in classification tasks~\cite{shorten2019survey,feng2021survey,zhang2021weakly}. In computer vision (CV),~\cite{krizhevsky2012imagenet,chatfield2014return,szegedy2015going} adopt strategies like flipping, cropping, tilting to perform DA.
In natural language processing (NLP),~\cite{xie2017data,coulombe2018text,niu2018adversarial,wei2019eda} find that native augmentation skills such as spelling errors, synonym replacement, deleting and swapping, can bring considerable performance improvement.
All these methods use various transformations for data augmentation, but they do not achieve equal success in different NLP tasks~\cite{yang2020g}. Sometimes, they fail to guarantee semantic consistency, and may even bring semantic errors that are harmful to classification. The reason lies in that data augmentation for NLP is in discrete space, so it can easily incur large deviation of semantics (\textit{e.g.} in sentiment classification task, deleting emotional words from a sentence will make its meaning completely different). 

Generally, given the size of generated data, their \emph{diversity} and \emph{quality} are crucial to the performance of targeted tasks~\cite{ash2019deep}.
Recent works have begun to emphasize the diversity or quality of augmented data. For example, in CV, AA~\cite{cubuk2019autoaugment}, Fast-AA~\cite{lim2019fast} and LTA~\cite{luo2020learn} employ agent networks to learn how to enhance diversity.
In NLP, language models are widely used to control generation quality, including Back-translation~\cite{sennrich2016improving,yu2018qanet}, Seq2seq models ~\cite{kobayashi2018contextual,kumar2019submodular,yang2020g}, GPT-2 ~\cite{radford2019language,anaby2020not,quteineh2020textual,liu2020data} and T5 ~\cite{dong2021data}. In addition, some works~\cite{morris2020textattack} in NLP utilize adversarial augmentation to enrich the diversity of the samples.
However, to the best of our knowledge, most existing works consider only either the quality or the diversity of augmented data, so cannot fully exploit the potential of data augmentation for NLP tasks.
Besides, recent existing DA methods for NLP tasks usually resort to pre-trained language models, are extremely inefficient due to huge model complexity and tedious finetuning, which limits the scope of their applications.

\begin{figure}
	\begin{center}
		\includegraphics[width=\linewidth]{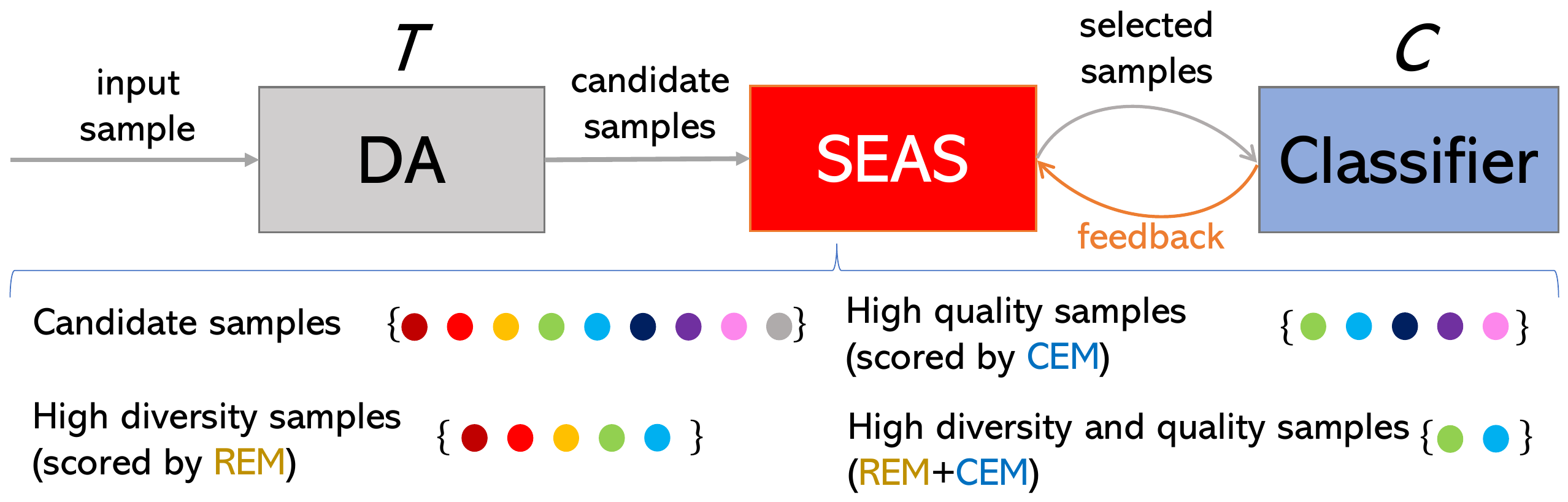}
	\end{center}
	\caption{The pipeline of EPiDA. EPiDA first evaluates the candidate samples from the perspectives of diversity and quality, then takes the samples of high diversity and quality as training samples. It can work with various DA algorithms and classification models.}
	\label{fig:intro}
\end{figure}

In this paper, we propose a new data augmentation framework for text classification. This framework is called EPiDA (the abbreviation of \underline{E}asy \underline{P}lug-\underline{i}n \underline{D}ata \underline{A}ugmentation), which employs two mechanisms to control the diversity and quality of augmented data: \emph{relative entropy maximization} (REM) and \emph{conditional entropy minimization} (CEM), where the former is for boosting diversity while the latter for ensuring quality.
Fig.~\ref{fig:intro} shows the pipeline of EPiDA. EPiDA consists of a DA algorithm, a classifier, and a \textit{Sample Evaluation And Selection} (SEAS) module. SEAS works with the DA algorithm and the classifier, and evaluates the candidate samples with the feedback of the classifier. With REM and CEM, SEAS can select samples of high diversity and quality to train the classifier continuously until the model converges.

The main contributions of this paper are as follows:
\begin{enumerate}
  \item We propose an easy plug-in data augmentation framework EPiDA for text classification. EPiDA can work with various existing DA algorithms and classification models, it is general, efficient, and easy-to-deploy.
  \item We design two mechanisms \emph{relative entropy maximization} (REM) and \emph{conditional entropy minimization} (CEM) to boost the diversity and quality of augmented data simultaneously in an explicit and controllable way.
  \item We conduct extensive experiments to evaluate EPiDA. Experimental results show that EPiDA outperforms existing DA methods, and works well with different DA algorithms and classification models.
\end{enumerate}
The rest of this paper is organized as follows: Sec.~\ref{sec:related-work} reviews related work and highlights the differences between our work and major existing methods. Sec.~\ref{sec:method} introduce our method in details. Sec.~\ref{sec:performance} presents the results of performance evaluation, and Sec.~\ref{sec:conclusion} concludes the paper.

\section{Related Work}\label{sec:related-work}
In this section, we first review the related work of DA for NLP, then expound the differences between our method and the major existing ones. According to the methodology of data generation, existing methods can be categorized into three types: rule-based, interpolation-based, and model-based, respectively.

\subsection{Rule-Based Methods}
These works use easy and predetermined transformations without model components. \cite{kolomiyets2011model,zhang2015character,wang2015s} use synonyms to replace words. EDA~\cite{wei2019eda} and AEDA~\cite{karimi2021aeda} introduce random insertions, swaps, and deletions. \citet{xie2017data} employed spelling errors to augment sentences. \citet{csahin2018data} conducted sentence rotating via dependency tree morphing. \citet{wei2021text} proposed a multi-task view of DA.
SUB$^2$~\cite{shi2021substructure} generates new examples by substituting substructures via constituency parse trees. Although these methods are easy to implement, they do not consider controlling data quality and diversity.

\subsection{Interpolation-Based Methods}
MIXUP~\cite{zhang2017mixup} pioneers this type of works by interpolating the input and labels of two or more real examples. Recently, many MIXUP strategies~\cite{verma2019manifold,yun2019cutmix} were proposed in CV. Due to the discrete nature of inputs of NLP tasks, such methods can be applied to NLP tasks only via padding and mixing embeddings or higher hidden layers~\cite{chen2020mixtext,si2021better}.

\subsection{Model-Based Methods}
Seq2seq and language models have been used to generate high quality samples. Among these approaches, Back-translation~\cite{sennrich2016improving,yu2018qanet} translates sentences into another language and then translates it back to the original language. RNNs and transformers are used to reconstruct sub-parts of real data with contextual information~\cite{kobayashi2018contextual,gao2019soft,yang2020g}.
Recently, we have witnessed the great success of large-scale pre-trained language models (PLMs) such as BERT~\cite{devlin2019bert}, XLNet~\cite{yang2019xlnet}, GPT-2~\cite{radford2019language} in NLP tasks. These state-of-the-art PLMs are also widely used to augment samples~\cite{ng2020ssmba,nie2020named,anaby2020not,quteineh2020textual,liu2020data,dong2021data}.
For example,
DataBoost~\cite{liu2020data} develops a reinforcement learning strategy to guide the conditional generation without changing the architecture of GPT-2. Besides, adversarial augmentation (\textit{i.e.,} attack, GANs) are also used to enrich the diversity of the generated samples~\cite{morris2020textattack,simoncini2021seqattack}.
Although model-based methods can control generation quality well via PLMs, they are computationally inefficient, which limits their applications.
\begin{table}
	\centering
	\resizebox{0.45\textwidth}{!}{
	\begin{tabular}{c|cccc}
		\toprule
		Method & Div  & Qua & LM & FB \cr
		\midrule
		AA~\cite{cubuk2019autoaugment} & $\checkmark$ &$\times$ &$\checkmark$ &$\checkmark$ \\
		EDA~\cite{wei2019eda} & $\checkmark$ &$\times$ &$\times$ &$\times$ \\
		DataBoost~\cite{liu2020data} & $\times$ &$\checkmark$ &$\checkmark$ &$\times$\\
		LearnDA~\cite{zuo2021learnda}  & $\checkmark$ & $\checkmark$ & $\checkmark$  &$\times$ \\
		VDA~\cite{zhou2021virtual} & $\checkmark$ & $\checkmark$ & $\checkmark$  &$\times$\\
		Ours EPiDA & $\checkmark$ & $\checkmark$ &$\times$ & $\checkmark$\\
		\bottomrule
	\end{tabular}
	}
	\caption{A qualitative comparison between EPiDA and major existing methods from four aspects: whether controlling the diversity and quality of the augmented data, whether using language model or agent network and whether using the feedback of the classifier.}
	\label{tab-diff}
\end{table}

\subsection{Differences between EPiDA and Existing Methods}
To expound the differences between EPiDA and typical existing methods, in Tab.~\ref{tab-diff} we present a qualitative comparison from four dimensions: whether controlling the diversity and quality of the augmented data, whether using pre-trained model (language model or agent network), and whether using the feedback from the classifier.

As shown in Tab.~\ref{tab-diff}, among the existing methods, most control only either diversity (\textit{e.g.} AA and EDA) or quality (\textit{e.g.} DataBoost) of augmented data, thus cannot completely leverage the potential of data augmentation. And most use language model or agent network, which is beneficial to data quality but also inefficient. Only the recent LearnDA~\cite{zuo2021learnda} and VDA~\cite{zhou2021virtual} consider both diversity and quality, and only AA uses feedback of the classifier. Our EPiDA addresses both diversity and quality of augmented data via the feedback of the classifier in an explicit and controllable way, without the help of any additional model components, which makes it not only more effective but also more efficient.

Note that in addition to the differences listed in Tab.~\ref{tab-diff}, our method EPiDA differs from LearnDA in at least three other aspects: 1) LearnDA employs perplexity score (PPL) and cosine similarity to measure diversity and quality respectively, while EPiDA adopts two mechanisms \emph{relative entropy maximization} (REM) and \emph{conditional entropy minimization} (CEM) to control diversity and quality, which is theoretically more rational and solid. 2) LearnDA is for event causality identification, while EPiDA is mainly for text classification. 3) LearnDA needs knowledge guidance, while EPiDA does not. These make it difficult to evaluate LearnDA in our experimental settings. Thus, we do not conduct performance comparison between EPiDA and LearnDA. Nevertheless, in our ablation study, we replace REM and CEM with PPL and cosine similarity in EPiDA, and our experimental results show that EPiDA with REM and CEM performs better than that with PPL and cosine similarity.
Besides, comparing with VDA that requires PLM to provide substitution probability, EPiDA is free of PLMs, and is more effective, efficient and practical.


\section{Method}\label{sec:method}

As shown in Fig.~\ref{fig:intro}, EPiDA consists of three components: a DA algorithm $T$, a classifier or classification model $C$, and a \textit{Sample Evaluation and Selection} (SEAS) module that is the core component of EPiDA. Generally, the DA algorithm and the classifier can be any of existing DA algorithms and classifiers. With the feedback of the classifier, SEAS evaluates candidate samples generated by the DA algorithm in terms of \emph{diversity} and \emph{quality} via the \textit{Relative Entropy Maximization} (REM) mechanism and the \textit{Conditional Entropy Minimization} (CEM) mechanism, and outputs the qualified samples to further train the classifier. 
So EPiDA can serve as a plug-in component to boost existing DA algorithms for training better target models. 

\subsection{The Rationale to Control DA}
\label{sec-motivation}
Consider a classification task with a dataset $X$ of $n$ samples: $X$=$\{(x_1,y_1)$,$(x_2,y_2)$,...,$(x_n,y_n)\}$. Here, $x_i$ is a sample, $y_i$ is its label. The loss function is
\begin{equation}
	\label{eq-aug}
	\begin{split}
		L_{o}(\omega)=\frac{1}{n}\sum_{i=1}^{n}l(\omega^\top \phi(x_i);y_i).
	\end{split}	
\end{equation}
%
where $\phi:\mathbb{R}^d \rightarrow \mathbb{R}^D$ is a finite-dimensional feature map, $\omega \in \mathbb{R}^D$ means learnable parameters, and $l$ can be a common loss function like
cross-entropy. 

Now we employ a DA algorithm $T$ to conduct augmentation for each sample in $X$. Let $t_i^j$ be the $j$-th sample generated by $T$ with $x_i$ as input, and $m$ samples are generated from $x_i$, the loss function for the generated samples can be written as
\begin{equation}
	\label{eq-aug-epida}
	\begin{split}
		L_{g}(\omega)=\frac{1}{n}\sum_{i=1}^{n}\frac{1}{m}\sum_{j=1}^{m}l(\omega^\top \phi(t_i^j);y_i).
	\end{split}	
\end{equation}
Here, 
we assume 1) $t_i^j$ and $x_i$ have the same label $y_i$, so we can use $y_i$ to optimize the new loss function; 2) Data augmentation does not significantly change the feature map $\phi$, that is, augmentation can maintain semantic consistency of the sample space. 
Now we combine the augmented samples into the original samples, thus the total loss function of EPiDA can be written as follows:
\begin{equation}
	\label{eq-EPiDA-tot}
	\begin{split}
		L(\omega)& = L_{o}(\omega) + L_{g}(\omega).
	\end{split}
\end{equation}

Recall that we use the feedback of the classifier $C$ to select samples. Specifically, we use the original training samples $X$ to pre-train the classifier $C$, and for each generated sample $t_i^j$, the feedback signal about $t_i^j$ from the classifier is used for evaluating $t_i^j$. When the generation process is over, all generated samples $\{t_i^j\}$ are used to train $C$ again.

First, we consider how to generate samples of high diversity. Intuitively, generated samples should be different from the original samples. Recalling that the classifier $C$ is pretrained by $X$, so for generated sample $t_i^j$, its loss $l(\omega^\top \phi(t_i^j);y_i)$ should be large. In this sense, given the classifier $C$ ($\omega$ is fixed), we select samples that meet the following objective function:
%
\begin{equation}
	\label{eq-adv-train}
	\begin{split}
		\mathop{\max}_{t_i^j}L_{g}(\omega,\phi(t_i^j)),
	\end{split}	
\end{equation}
which means that we are to generate ``hard'' samples for the classifier to cope with. 

Second, we consider how to control the quality of augmented data. Recall that we assume for each augmented sample $t_i^j$, its label $y_i$ keeps unchanged, so we can use the original label to evaluate the loss function. However, due to the discrete nature of language, it is nontrivial for augmented samples to meet this assumption. Taking the sentiment analysis task for example, suppose we use EDA~\cite{wei2019eda} to augment $x_i$=``you'll probably love it'', EDA may delete the word ``love''. Obviously, the resulting sentence breaks the semantic consistency. 
To guarantee semantic consistency, we limit the semantic deviation of $\phi(t_i^j)$ from $\phi(x_i)$. Let $M$ and $\rho$ be a metric function to measure semantic difference between samples and a threshold respectively, we impose the following constraint on $\phi(t_i^j)$:
\begin{equation}
	\label{eq-cons}
	\begin{split}
		|M(\omega^\top \phi(t_i^j),\omega^\top \phi(x_i))| \le \rho.
	\end{split}	
\end{equation}

Thus, we can enhance data diversity by optimizing Eq.~(\ref{eq-adv-train}), and improve data quality using Eq.~(\ref{eq-cons}). The problem turns to solve Eq.~(\ref{eq-adv-train}) and Eq.~(\ref{eq-cons}).

\subsection{Relative Entropy Maximization}
\label{sec-rem}
We rewrite the objective function in Eq.~(\ref{eq-adv-train}) via:
{\small
\begin{equation}
	\label{eq-gm-proof}
	\begin{split}
		&L_{g}(\omega,\phi(t_i^j)) = \frac{1}{nm}\sum_{i=1}^{n} \sum_{j=1}^{m} l(\omega^\top \phi(t_i^j);y_i) \\ &= \frac{1}{nm}\sum_{i=1}^{n} \sum_{j=1}^{m} D(p(\omega^\top \phi(t_i^j)),p(y_i)) + H(p(y_i)),
	\end{split}
\end{equation}}
where $p$, $H$, $D$ indicate probability distribution, \textit{Shannon entropy}, and \textit{relative entropy} respectively, and actually $H(p(y_i))=0$ since $p(y_i)$ is a one-hot vector. According to Eq.~(\ref{eq-gm-proof}), we try to augment samples with large relative entropy under the given labels. Thus, we call this method \textit{relative entropy maximization} (REM) mechanism. 
As relative entropy measures the difference between the two distributions $p(\omega^\top \phi(t_i^j))$ and $p(y_i)$, the larger the difference is, the more diverse the augmented sample is. Therefore, we define the \emph{diversity score} $s_{div}^{ij}$ of augmented sample $t_i^j$ as follows:
\begin{equation}
	\label{eq-gm-score}
	\begin{split}
		s_{div}^{ij}=D(p(\omega^\top \phi(t_i^j)),p(y_i)).
	\end{split}	
\end{equation}

\subsection{Conditional Entropy Minimization}

We use \textit{conditional entropy} as the metric function in Eq.~(\ref{eq-cons}) to constrain the semantic deviation of $\phi(t_i^j)$ from $\phi(x_i)$, \textit{i.e.},  $M(\cdot,\cdot):=H(\cdot|\cdot)$. Then, Eq.~(\ref{eq-cons}) can be rewritten to
\begin{equation}
	\label{eq-ce}
	\begin{split}
		H(p(\omega^\top \phi(t_i^j))|p(\omega^\top \phi(x_i))) \le \rho.
	\end{split}
\end{equation}
%
where $H(\cdot,\cdot)$ is \emph{conditional entropy}. Furthermore, to meet Eq.~(\ref{eq-ce}), we select samples $\{t_i^j\}$ by solving the following optimization problem:
\begin{equation}
	\label{eq-cem}
	\begin{split}
		\mathop{\min}_{t_i^j} H(p(\omega^\top \phi(t_i^j))|p(\omega^\top \phi(x_i))).
	\end{split}
\end{equation}
\noindent We call this \textit{conditional entropy minimization} (CEM) mechanism.
The smallest value of $H(p(\omega^\top \phi(t_i^j))|p(\omega^\top \phi(x_i)))$ is 0, indicating that given $p(\omega^\top \phi(x_i))$, $p(\omega^\top \phi(t_i^j))$ is exactly predictable. Eq.~(\ref{eq-cem}) can also be expanded to the difference between Shannon entropy $H$ and mutual information $I$, \textit{i.e.}, $H(X|Y)$=$H(X)$-$I(X,Y)$. In other words, CEM minimizes the entropy of the selected sample $t_i^j$ and maximizes the mutual information between $t_i^j$ and the original sample $x_i$, which means that CEM tries to augment samples of high prediction probability and high similarity with the original sample. As in REM, we define the \emph{quality score} $s_{qua}^{ij}$ of augmented sample $t_i^j$ as
\begin{equation}
	\label{eq-cem-score}
	\begin{split}
		s_{qua}^{ij}=-H(p(\omega^\top \phi(t_i^j))|p(\omega^\top \phi(x_i))).
	\end{split}	
\end{equation}
\label{sec-mim}

\begin{algorithm}[!h]
	\SetAlgoLined
	\KwIn{Classification model $\omega$, input sample $x_i$ and its label $y_i$, DA algorithm $T$, augmentation number $m$ and amplification factor $K$.}
	\KwOut{$m$ augmented samples.}
	\begin{small}
	// assign the set $T(x_i)$ of $mK$ candidates to array $\hat{t}_i$
	 \end{small}\\
	$\hat{t}_i \leftarrow T(x_i)$\;
	$s_{div} ,s_{qua}, s_{tot}= \mathbb{R}^{K*m}, \mathbb{R}^{K*m}, \mathbb{R}^{K*m}$\;
	\For{$j=1,2,\ldots,K*m$}{
		Calculate $s_{div}^{ij}$ via Eq.~(\ref{eq-gm-score}) \;
		Calculate $s_{qua}^{ij}$ via Eq.~(\ref{eq-cem-score}) \;
	}
	Take Min\_Max\_Norm for $s_{div}$ and $s_{qua}$\;
	$s_{tot} = s_{div} + s_{qua}$ \;
	\begin{small}
	// find the subscripts of the top $m$ small elements
	\end{small}
	\\
	$id = \mathop{\mathrm{argtopm}}(-s_{tot})$ \;
	Return $\hat{t}_i[id]$ \;
	\caption{EPiDA Data Augmentation.}
	\label{algo-epda}
\end{algorithm}

\subsection{Algorithm and Implementation}
The procedure of EPiDA is presented in Alg.~\ref{algo-epda}. For each input sample $x_i$, EPiDA outputs $m$ augmented samples. First, we employ $T$ to generate $K*m$ candidate augmented samples for $x_i$, where $K$ is a hyperparameter to amplify the number of candidate samples, which is called \emph{amplification factor}. Then, for each augmented sample, we use REM and CEM to evaluate its diversity score ($s_{div}$) and quality score ($s_{qua}$), respectively. Next, we adopt Min\_Max\_Norm to make $s_{div}$ and $s_{qua}$ fall in [0,1]. After that, we add them together as the overall score of the sample, and sort all the augmented samples in descending order according to their scores. Finally, we take the top $m$ samples from all the $K*m$ candidate samples as the output, and utilize them to train the classifier.

\begin{table}
\resizebox{0.45\textwidth}{!}{
	\centering
	\begin{tabular}{c|c@{\hspace{0.4em}}c@{\hspace{0.4em}}c@{\hspace{0.4em}}}
		\toprule
		Sentence & $s_{div}$ & $s_{qua}$ & $s_{tot}$  \cr
		\midrule
		\tabincell{c}{Go Set a Watchman comes out Tues-\\day and I'm really \underline{excited} for it}& 0.00 & 1.00 &1.00 \\
		\midrule
		\tabincell{c}{Go Set a Watchman comes out Tues-\\day and I'm really \underline{mad} for it}& 0.96 & 0.03 &0.99 \\
		\tabincell{c}{Go Set a Watchman out Tuesday \\and I'm really \underline{excited} for it}& 0.05 & 0.92 &0.97 \\
		\tabincell{c}{Go Set a security guard comes out Tues \\and I'm really \underline{excited} for it}& 0.86 & 0.15 &1.01 \\
		\bottomrule
	\end{tabular}
	}
	\caption{An example from the sentiment analysis task to demonstrate the effect of REM and CEM. The first row is the original sample, the rest are the augmented samples. Underlined are salient words.}
	\label{tab-example}
\end{table}

By nature, the goals of REM and CEM are conflicting, i.e., a sample of high diversity is more probably of low quality, and vice versa.
We give an example in Tab.~\ref{tab-example} to demonstrate this point. REM encourages to change salient words, which is prone to break the semantic consistency (see the 3rd row, ``excited'' is changed to ``mad'', leading to large diversity score but small quality score). However, CEM tends to make the augmented samples keep semantic consistency, i.e., has large quality score but small diversity score (see the 4th row, ``comes'' is deleted). By jointly considering REM and CEM, satisfactory samples with balanced diversity and quality can be found (see the 5th row).

Besides, the calculation of $s_{div}$ and $s_{qua}$ requires the feedback of the classifier, so we first pre-train the classifier using the original samples, then with EPiDA we can generate samples of high diversity and quality for the classifier continuously.
\label{sec-epda}

\begingroup
\setlength{\tabcolsep}{3pt}
\begin{table*}[!t]
	\centering
	\resizebox{0.9\textwidth}{!}{%
		\begin{tabular}{@{}cccccccccc@{}}
			\toprule
			\multirow{2}{*}{\textbf{Method}} & \multicolumn{3}{c}{\textbf{Sentiment}} & \multicolumn{3}{c}{\textbf{Irony}} & \multicolumn{3}{c}{\textbf{Offense}} \\ \cmidrule(l){2-10}
			& 10\% & 40\% & PPL & \multicolumn{1}{l}{10\%} & 40\% & PPL & \multicolumn{1}{l}{10\%} & 40\% & PPL \\ \midrule
			\begin{tabular}[c]{@{}c@{}}\textbf{EDA}\\ (randomly delete, swap etc.)\end{tabular} & 0.560 & 0.608 & 41.22 & 0.530 & 0.515 & 76.07 & 0.637 & 0.629 & 37.37\\ \midrule
			\begin{tabular}[c]{@{}c@{}}\textbf{Contextual Word Embs Aug.}\\ (insert, replace using Bi-RNN LM)\end{tabular} & 0.610 & 0.627 & 1043.18 & 0.518 & 0.593 & 1146.40 & 0.663 & 0.713 & 1729.62\\ \midrule
			\begin{tabular}[c]{@{}c@{}}\textbf{Back-translation Aug.}\\ (Eng. $\rightarrow$ Fr. $\rightarrow$ Eng. as aug. text)\end{tabular} & 0.617 & 0.620 & 474.29 & 0.520 & 0.541 & 423.32 & 0.655 & 0.724 & 345.23\\ \midrule
			\begin{tabular}[c]{@{}c@{}}\textbf{Data Boost}\\ (RL-guided conditional generation)\end{tabular}  & 0.591 & 0.642 & 56.23 & 0.591 & 0.639 & 77.40 &\textbf{0.695} & \textbf{0.784} & 35.18 \\ \midrule
			\begin{tabular}[c]{@{}c@{}}\textbf{Ours EPiDA: REM only}\\ (EDA as the DA algorithm)      \end{tabular} & 0.619 & 0.650 & 66.86 & 0.624 & 0.665 & 77.09 & 0.662 & 0.673 & 81.13
			\\ \midrule
			\begin{tabular}[c]{@{}c@{}}\textbf{Ours EPiDA: CEM only}\\ (EDA as the DA algorithm)      \end{tabular} & 0.629 & 0.659 & \textbf{8.10} & 0.629 & 0.666 & \textbf{12.11} & 0.668 & 0.670 & \textbf{8.57}
			\\ \midrule
			\begin{tabular}[c]{@{}c@{}}\textbf{Ours EPiDA: REM+CEM}\\ (EDA as the DA algorithm)      \end{tabular} & \textbf{0.639} & \textbf{0.659} & \textbf{25.40} & \textbf{0.651} & \textbf{0.687} & \textbf{53.17} & 0.680 & 0.687 & \textbf{32.56}
			\\ \bottomrule
		\end{tabular}%
		
	}
	\caption{Performance comparison with existing augmentation methods. 10\%: 10\% original data + 30\% augmented data ($m$ = 3); 40\%: 40\% original data + 40\% augmented data ($m$ = 1). We report the F1 score of the BERT classifier averaged over five repeated experiments on each dataset. We also report the perplexity score (PPL) of 10,000 randomly sampled data augmented by each method, where PPL is evaluated by the kenLM language model trained on the training data of each task.}
	\label{tab:major}
\end{table*}
\endgroup

\section{Performance Evaluation}\label{sec:performance}
In this section, we conduct extensive experiments to evaluate EPiDA, including performance comparison with SOTA methods, performance evaluation when working with different DA algorithms and classification models, ablation study, and qualitative visualization of samples augmented by EPiDA.

\subsection{Datasets and Settings}
Datasets for five different tasks are used in our experiments: \emph{Question Classification}
~\cite{li2002learning}
 (\textbf{TREC}, $N$=5,452), \emph{News Classification}
 ~\cite{zhang2015character} (\textbf{AGNews}, $N$=120,000), \emph{Tweets Sentiment Analysis}
 ~\cite{rosenthal2017semeval} (\textbf{Sentiment}, $N$=20,631), \emph{Tweets Irony Classification}
 ~\cite{van2018semeval} (\textbf{Irony}, $N$=3,817) and \emph{Tweets Offense Detection}
 ~\cite{founta2018large} (\textbf{Offense}, $N$=99,603), where $N$ is the number of training samples. To fully demonstrate the performance of data augmentation, we use only part of each dataset. In the following experiments, the percentage (\%) that follows the task name means the ratio of training data used from each dataset, \textit{e.g.} Irony 1\% means that 1\% of the dataset is used. Macro-F1 (F1 for binary tasks) is used as performance metric, and all the experiments are repeated five times. The amplification factor $K$ is set to 3.

\begingroup
\setlength{\tabcolsep}{3pt}
\begin{table*}[!t]
	\centering
	\resizebox{0.9\textwidth}{!}{%
		\begin{tabular}{@{}ccccccccccc@{}}
			\toprule
			\multirow{2}{*}{\textbf{Method}} & \multicolumn{2}{c}{\textbf{TREC}} & \multicolumn{2}{c}{\textbf{AGNews}} & \multicolumn{2}{c}{\textbf{Sentiment}} & \multicolumn{2}{c}{\textbf{Irony}} &
			\multicolumn{2}{c}{\textbf{Offense}}\\ \cmidrule(l){2-11}
			& 1\% & 10\% & 0.05\% & 0.1\%  & 1\% & 10\%  & 1\% & 10\% & 0.1\% & 1\%\\ \midrule
			\textbf{CNN} &0.722 & 0.806 & 0.745 & 0.826 & 0.446 & 0.584 &0.534 &0.616 &0.479 &0.548\\
			+\textit{EPiDA with EDA} &\textbf{0.745} & 0.814 & 0.806 & \textbf{0.829} & 0.520 & 0.598 &\textbf{0.579} &0.621 &\textbf{0.493} &0.551\\
			+\textit{EPiDA with CWE} &0.737 & 0.817 & 0.806 & 0.826 & 0.524 & 0.598 &0.578 &0.623 &0.488 &\textbf{0.556}\\
			+\textit{EPiDA with TextAttack} &0.723 & \textbf{0.838} & \textbf{0.819} & \textbf{0.829} & \textbf{0.527} & \textbf{0.600} & 0.568 &\textbf{0.631} &0.481 &0.549\\
			\midrule
			\textbf{BERT} &0.769 & 0.914 & 0.759 & 0.820 & 0.507 & 0.602 &0.589 &0.635 &0.508 &\textbf{0.630}\\
			+\textit{EPiDA with EDA} &\textbf{0.786} & \textbf{0.931} & 0.813 & 0.832 & 0.538 & 0.641 &\textbf{0.598} &0.652 &\textbf{0.525} &0.629\\
			+\textit{EPiDA with CWE} &0.780 & 0.922 & \textbf{0.821} & 0.834 & 0.546 & \textbf{0.642} &0.597	 &\textbf{0.655} &0.517 &\textbf{0.630}\\
			+\textit{EPiDA with TextAttack} &0.762 & 0.930 & 0.816 & \textbf{0.839} & \textbf{0.551} & 0.621 &0.595 &0.634 &0.505 &0.626\\
			\midrule
			\textbf{XLNet} &0.746 &0.904  &0.749 &0.776 &\textbf{0.563} &\textbf{0.620} &0.576 &0.651 &0.522 &0.626\\
			+\textit{EPiDA with EDA} &0.756 &0.906 &0.768 &0.790 &0.556 &0.609 &0.588 &0.651 &\textbf{0.536} &0.625 \\
			+\textit{EPiDA with CWE} &0.750 &0.894 &0.779 &0.795 &0.554 &0.608 &\textbf{0.592} &\textbf{0.659} &0.532 &\textbf{0.627}\\
			+\textit{EPiDA with TextAttack} &\textbf{0.758} & \textbf{0.909} & \textbf{0.790} & \textbf{0.798} & 0.555 & 0.591 &0.570 &0.654 &0.528 &0.618\\
			\bottomrule
		\end{tabular}%
	}
	\caption{Performance comparison with different DA algorithms and classification models on five classification tasks.}
	\label{tab:generality}
\end{table*}
\endgroup

\subsection{Comparing with SOTA Methods}
Here we carry out performance comparison with major SOTA methods to show the superiority of EPiDA on three datasets: \textbf{Sentiment}, \textbf{Irony} and \textbf{Offense}.
For a fair comparison, we strictly follow the experimental setting of DataBoost~\cite{liu2020data}: we do only one round of augmentation to ensure that the number of samples of our method is consistent with that of the other methods, and use BERT as the classifier. We use the widely used EDA as the DA algorithm of EPiDA.
We do not use DataBoost because it is not yet open-sourced.
The experimental results are presented in Tab.~\ref{tab:major}.

From Tab.~\ref{tab:major}, we can see that 1) with the help of EPiDA, the performance of EDA is greatly improved. In particular, comparing with the original EDA, EPiDA gets performance improvement of 14.1\%, 8.39\%, 22.83\%, 33.40\%, 6.75\% and 9.22\% in six task settings, respectively. 2) Our method outperforms DataBoost in four settings. In particular, EPiDA$+$EDA gets performance improvement of 8.12\%, 2.65\%, 10.15\% and 6.99\% in various settings of the \textbf{Sentiment} and \textbf{Irony} tasks. 3) The variants of EPiDA that utilize only REM or CEM to enhance diversity or quality are inferior to using both, which demonstrates the effectiveness of joint enhancement. 4) DataBoost performs better in the \textbf{Offense} task, the reason lies in that DataBoost can create novel sentences from \textbf{Offense} (a relatively huge corpus) via GPT-2, while EDA only conducts word-level augmentation, which limits EPiDA's performance. 5) We also present PPL as an auxiliary metric to measure the generation perplexity. Our method outperforms the others due to the high quality of data generation.
We also provide experimental comparisons with other DA approaches (SUB$^2$ and VDA) and generation speed results in the supplementary file. In conclusion, EPiDA is a powerful and efficient technique.

\subsection{Performance with Different DA Algorithms and Classifiers}
EPiDA is a plug-in component that can work with different DA algorithms and classifiers. Here, to check how EPiDA performs with different DA algorithms and classifiers, we consider three frequently-used DA algorithms: rule-based EDA~\cite{wei2019eda}, model-based CWE~\cite{kobayashi2018contextual} and Attack-based TextAttack~\cite{morris2020textattack}, and three different classifiers: CNN~\cite{kim2014convolutional}, BERT~\cite{devlin2019bert} and XLNet~\cite{yang2019xlnet}.
And to show that EPiDA can cope with different NLP classification tasks, we present the results on five different tasks: \textbf{TREC}, \textbf{AGNews}, \textbf{Sentiment}, \textbf{Irony} and \textbf{Offense}. In order to fully evaluate the performance of DA algorithms, we use only a small part of the training data. The experimental results are presented in Tab.~\ref{tab:generality}. Here, we remove the restriction of only one time augmentation so that EPiDA can continuously generate qualified samples. We call this \emph{online augmentation}, to differentiate it from one-time augmentation.
As shown in Tab.~\ref{tab:generality}, EPiDA is applicable to various NLP classification tasks. Although these tasks have different forms of data (questions or tweets) and different degrees of classification difficulty, EPiDA boosts performance on these tasks in almost all cases. More details on how EPiDA controls the generation quality are discussed in ablation study. Besides, we can also see that EPiDA works well with the three DA algorithms EDA, CWE and TextAttack. All achieve improved performance in most cases.
For the three different classification models: CNN, BERT and XLNet, with the help of EPiDA, they all but XLNet on Sentiment get classification performance improvement, which shows that EPiDA is insensitive to classification models.

\subsection{Ablation Study}
\label{sec:exp:ab}
Here we conduct ablation study to check the effectiveness of different EPiDA configurations. We take CNN as the classifier, EDA as the DA algorithm and report the Macro-F1 score over five repeated experiments on \textbf{TREC} 1\% and \textbf{Irony} 1\%. Tab.~\ref{tab:abstudy} shows the experimental results.
\begin{table}
\resizebox{0.48\textwidth}{!}{
	\centering
	\begin{tabular}{c@{\hspace{0.4em}}|c@{\hspace{0.4em}}c@{\hspace{0.4em}}c@{\hspace{0.4em}}c@{\hspace{0.4em}}c@{\hspace{0.4em}}|cc}
		\toprule
		 ID&DA&REM & CEM   & OA  & PT & TREC 1\% & Irony 1\% \cr
		\midrule
		1&-&- & - & - &- & 0.722 & 0.534   \\ 
		2&$\checkmark$ & - & - & - & - & 0.736 & 0.474   \\ 
		3&$\checkmark$ & - & - & $\checkmark$ & - &0.723 & 0.550  \\ 
		4&$\checkmark$ & $\checkmark$ & - & - &$\checkmark$ & 0.729 & 0.557   \\ 
		5&$\checkmark$ & - & $\checkmark$ & - &$\checkmark$ & 0.723 & 0.559   \\ 
		6&$\checkmark$ & $\checkmark$ & $\checkmark$ & - &- &0.734  &0.548  \\ 
		7&$\checkmark$ & $\checkmark$ & $\checkmark$ & $\checkmark$ &- &0.739 &0.575  \\ 
		8&$\checkmark$ & $\checkmark$ & $\checkmark$ & -&$\checkmark$  &0.740 & 0.576  \\ 
		9&$\checkmark$ & $\checkmark$ & $\checkmark$ & $\checkmark$ &$\checkmark$ &0.745 &0.579  \\ 
		\bottomrule
	\end{tabular}
	}
	\caption{Ablation study on TREC 1\% and Irony 1\%. Here, DA, REM, CEM, OA and PT indicate whether applying DA, REM, CEM, online DA (performing multiple DA operations) and pre-training, respectively.}
	\label{tab:abstudy}
\end{table}


\textbf{Effect of REM and CEM}. The 4th and 5th rows show the results with only REM and CEM, respectively. Both of them perform better than the baseline (1st row), but not as good as the combined case (the 8th row). On \textbf{TREC} (relatively simple task), REM outperforms CEM (0.729 vs. 0.723), while on \textbf{Irony} (relatively hard task), CEM outperforms REM (0.559 vs. 0.557). Using only REM can limitedly boost performance since REM promotes the generation of high diversity samples, which may have wrong labels. And using only CEM is also not enough to fully tap the performance as CEM tends to generate redundant samples.

We also compare our `REM + CEM' with `PPL + cosine similarity' used in LDA~\cite{zuo2021learnda}. Our method achieves the performance of 0.740 and 0.576 on \textbf{TREC} 1\% and \textbf{Irony} 1\%, while the latter achieves 0.730 and 0.562. This shows that our `REM + CEM' is more effective.

\textbf{Effect of online augmentation}. Comparing the results of the 2nd and the 3rd rows, the 6th and the 7th rows, the 8th and the 9th rows, we can see that generally online augmentation can boost performance, as online augmentation can generate sufficient qualified samples to train the model. 

\textbf{Effect of pre-training}. As REM and CEM use the feedback of the classifier, a pre-trained classification model should be beneficial to REM and CEM. By comparing the results of the 6th and the 8th rows, the 7th and the 9th rows, it is obvious that pre-training can improve performance. 

\textbf{Effect of normalization}.
In Alg.~\ref{algo-epda}, we normalize $s_{div}$ and $s_{qua}$. Here, we check the effect of normalization. With the same experimental settings, the performance results on \textbf{TREC} 1\% and \textbf{Irony} 1\% without normalization are 0.732 and 0.568, lower than the normalized results 0.740 and 0.576. This shows that normalization is effective.

\textbf{How to combine REM and CEM}?
How to combine REM and CEM is actually how to combine the values of $s_{div}$ and $s_{qua}$. We consider three simple schemes: addition ($s_{tot}$ = $s_{div}$ + $s_{qua}$), multiplication ($s_{tot}$ = $s_{div} * s_{qua}$) and weighted addition ($s_{tot}$ = $\alpha s_{div}$ + $(1-\alpha)s_{qua}$, $\alpha$ is a hyperparameter to tradeoff REM and CEM). Note that for multiplication, there is possibly an extreme situation: after normalization, $s_{div}$ or $s_{qua}$ may be very small and even approaches 0, then the multiplication result is very small or even zero, which means that REM and CEM do not take effect in sample generation. In our experiments, the multiplication scheme achieves performance of 0.725 and 0.572 on \textbf{TREC} 1\% and \textbf{Irony} 1\%, lower than the addition scheme 0.740 and 0.576. As for weighted addition, we find that setting $\alpha = 0.5$ can achieve satisfactory results (see the supplementary file). This is actually equal to the addition scheme. Therefore, in our experiments, we use only the addition scheme.

\textbf{Quality and diversity metrics}.
Here, we provide another two metrics to verify EPiDA from the perspective of quality and diversity. For quality, we use the augmentation error rate. As for diversity, we calculate the average distance of samples before and after augmentation (ignoring wrong samples). From the perspective of quality and diversity, a good DA should has a \emph{small} error rate but a \emph{large} distance. Experimental results are given in Tab.~\ref{tab:error}. We can see that EPiDA gets better trade-off between error rate and distance.
\begin{table}[!t]
	\centering
	\resizebox{0.48\textwidth}{!}{
	\begin{tabular}{ccccc}
		\toprule
		Metric & EDA & REM only  & CEM only & EPiDA\cr
		\midrule
		Error Rate$\downarrow$  & 3.05\% & 6.75\%  & 0.64\% & 1.53\% \\
		Distance$\uparrow$($\times$ $10^{-2}$)  & 0.54 & 1.21 & 0.25 & 0.78 \\
		\bottomrule
	\end{tabular}
	}
	\caption{\emph{Error rate} and \emph{distance} on \textbf{Sentiment} dataset.}
	\label{tab:error}
\end{table}

\textbf{Effect of the amplification factor $K$}.
The amplification factor $K$ determines the size $Km$ of candidate samples from which $m$ samples are chosen. On the one hand, with a large $K$, we have more choices, which seems beneficial to diversity. On the other hand, more candidate samples make the selected samples more homogenous, not good for diversity.
By grid search, we set $K$ to 3 in our experiments, the experimental results are shown in the supplementary file.
\begin{figure}
	\begin{center}
		\includegraphics[width=\linewidth]{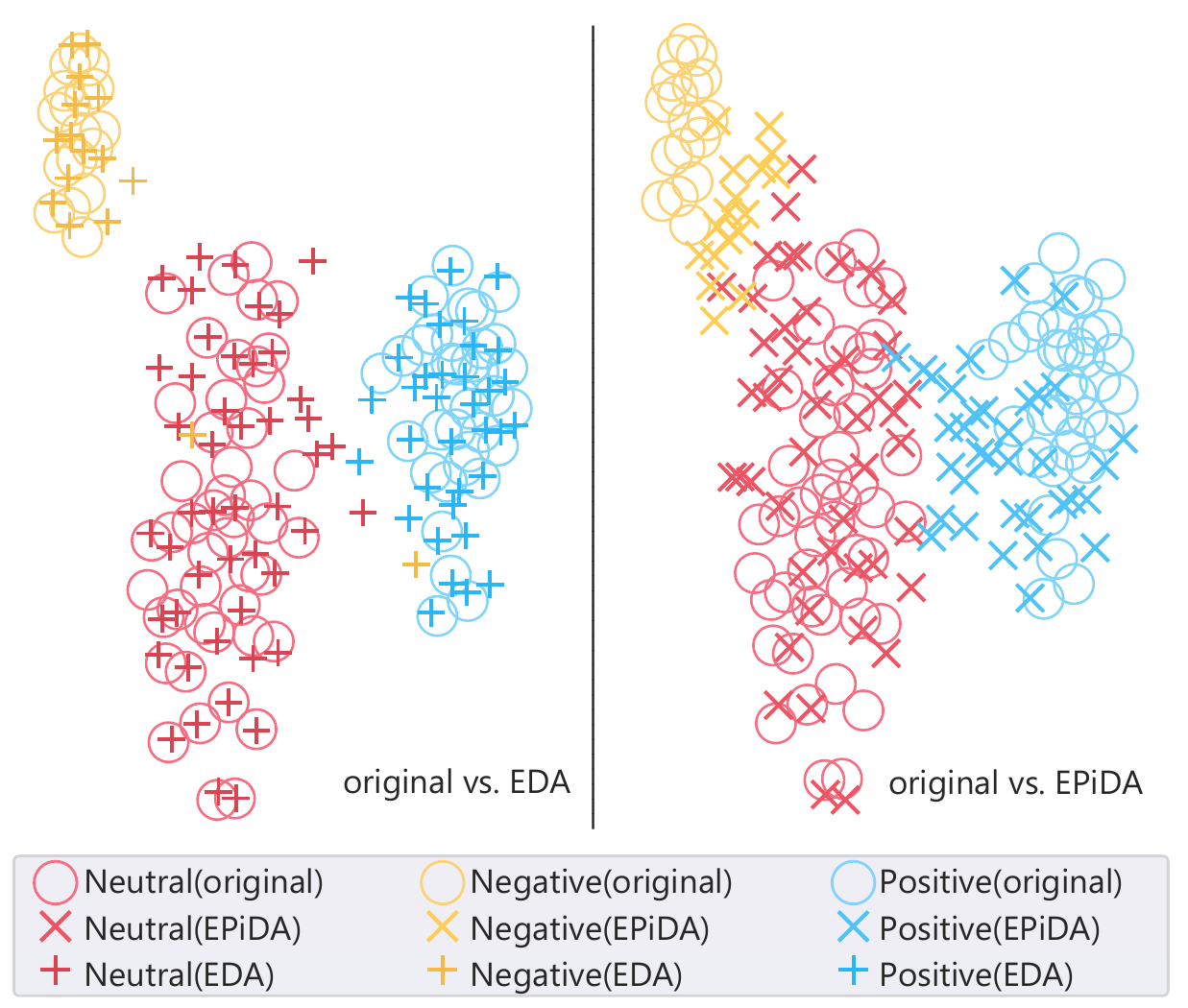}
	\end{center}
	\caption{Hidden state visualization of the sentiment analysis task by t-SNE. Left: original and EDA augmented data; Right: original and EPiDA augmented data. Different colors mean different classes (Neutral, Negative and Positive), and different shapes mean different samples (original, generated by EPiDA and EDA).}
	\label{fig:vis}
\end{figure}

\subsection{Visualization Effect of EPiDA}
Above we give comprehensive quantitative performance evaluation of EPiDA, here to intuitively illustrate the effectiveness of EPiDA, we visualize some augmented samples of EPiDA, and compare them with that of EDA. Specifically, we utilize BERT as the classifier and visualize its hidden state on the sentiment analysis task via t-SNE~\cite{van2008visualizing}.
Fig. ~\ref{fig:vis} shows the results.
In terms of data quality, we find that two negative samples generated by EDA are located in Neural and Positive classes, while samples generated by EPiDA are generally properly located. And in the point of view of diversity, samples generated by EPiDA extend the distributed areas of the original data, while samples generated by EDA are mainly located in the areas of the original samples. This shows that samples generated by EPiDA are more diverse than those generated by EDA.

\section{Conclusion}\label{sec:conclusion}
In this paper, we present an easy plug-in data augmentation technique EPiDA to control augmented data diversity and quality via two mechanisms: \textit{relative entropy maximization} and \textit{conditional entropy minimization}. Through extensive experiments, we show that EPiDA outperforms existing methods, and can work well with different DA algorithms and classification models. EPiDA is general, effective,  efficient, and easy-to-deploy. In the future, more
verification of our method is expected to be conducted on other classification tasks.

\section*{Acknowledgement}
This work was partially supported by National Key R\&D Program of China under grant No.~2021YFC3340302, and Alibaba Innovative Research (AIR) programme under contract No.~SCCW802020046613.

\bibliography{naacl-2022-cr}
\bibliographystyle{acl_natbib}
\appendix
\begin{figure}
	\begin{center}
		\includegraphics[width=\linewidth]{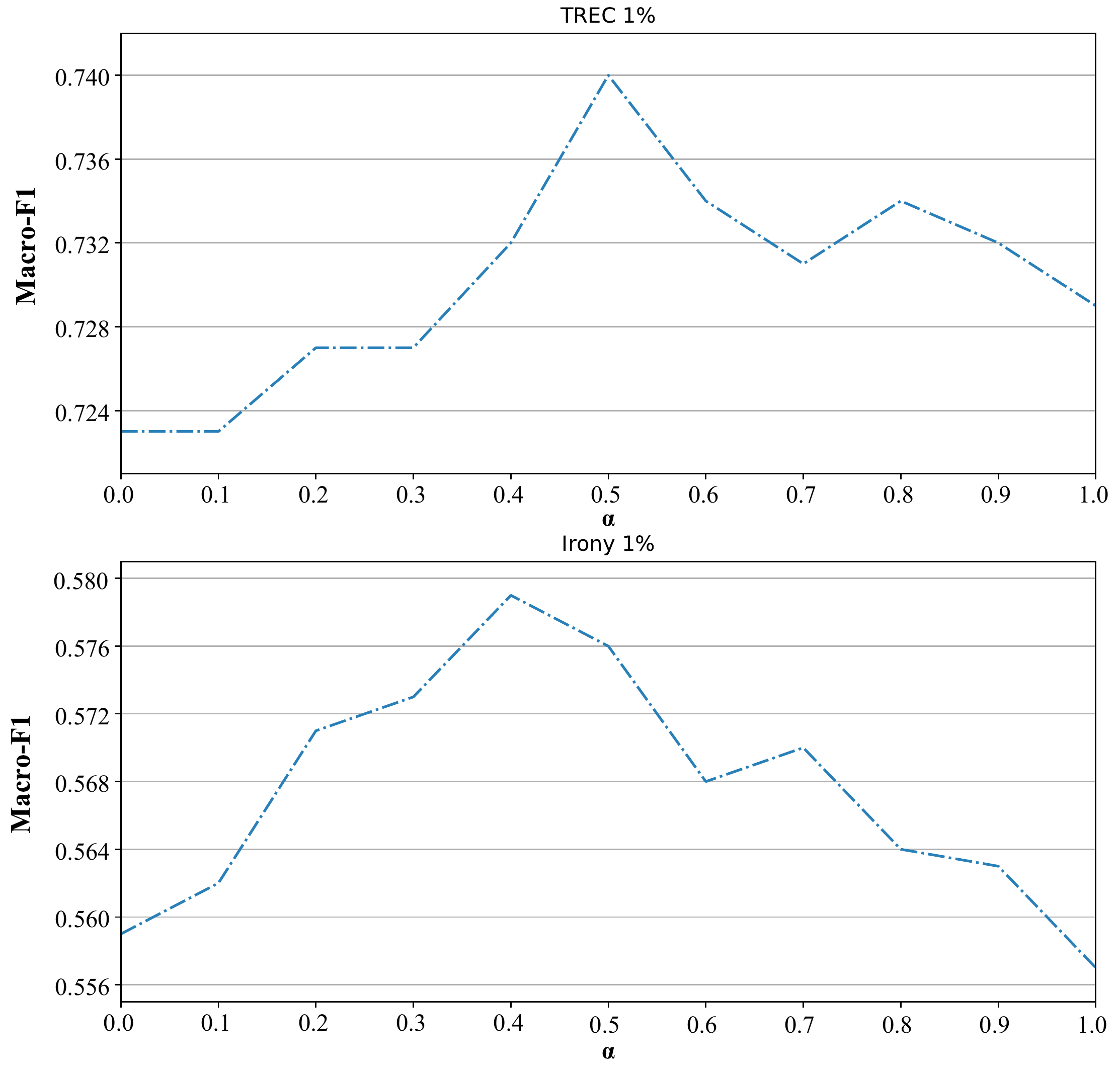}
	\end{center}
	\caption{The Macro-F1 classification performances under different $\alpha$. Top: \textbf{TREC} 1\%, Bottom: \textbf{Irony} 1\%.}
	\label{fig:supp:alpha}
\end{figure}
\section{Weighted Addition of REM and CEM}
Here we discuss the usage of weighted addition to combine REM and CEM. That is to say we introduce an additional hyperparameter $\alpha$, $\alpha \in[0,1]$ to control the trade-off of REM and CEM:
\begin{equation}
	\label{eq-trade-off}
	\begin{split}
		s_{tot} = \alpha s_{div} + (1-\alpha) s_{qua}
	\end{split}
\end{equation}
A larger $\alpha$ highlights diversity and suppresses quality, and vice versa. 

We discuss the influence of $\alpha$ on two datasets: \textbf{TREC} and \textbf{Irony}. We also use CNN~\cite{kim2014convolutional} as the classifier and Macro-F1 as the metric and report the average results over five times repeated experiments. The classification performance under different $\alpha$ is presented in Fig.~\ref{fig:supp:alpha}.

As shown in Fig.~\ref{fig:supp:alpha}, in \textbf{TREC} and \textbf{Irony} tasks, the best values of $\alpha$ are 0.5 and 0.4, respectively. Although $\alpha=0.4$ (0.579 vs. 0.576) performs better on the \textbf{Irony} task, 0.5 is sufficient to achieve satisfactory results on both tasks. Ergo,  we set $\alpha$ to 0.5 in this paper. 

\section{Ablation Study on Loss Function}
Here we take an ablation study to support the combined loss function used in our paper. Actually, they are three loss functions in this paper.

The first one is the original loss function without performing DA
\begin{equation}
	\label{eq-org}
	\begin{split}
		L_{o}(\omega)=\frac{1}{n}\sum_{i=1}^{n}l(\omega^\top \phi(x_i);y_i),
	\end{split}	
\end{equation}
which means we do not take DA to enrich the training data.

The second is the new loss function after using DA:
\begin{equation}
	\label{eq-aug}
	\begin{split}
		L_{g}(\omega)=\frac{1}{n}\sum_{i=1}^{n}\frac{1}{m}\sum_{j=1}^{m}l(\omega^\top \phi(t_i^j);y_i).
	\end{split}	
\end{equation}

The third is the combined loss function:
\begin{equation}
	\label{eq-EPiDA-tot}
	\begin{split}
		L(\omega)& = L_{o}(\omega) + L_{g}(\omega).
	\end{split}
\end{equation}
\begin{table}
	\centering
	\begin{tabular}{c|cc}
		\toprule
		Loss Function & TREC 1\% & Irony 1\% \cr
		\midrule
		Eq.~(\ref{eq-org}) & 0.722 & 0.534   \\ 
		Eq.~(\ref{eq-aug}) & 0.736 & 0.474   \\ 
		Eq.~(\ref{eq-EPiDA-tot})  &0.740 & 0.576  \\ 
		\bottomrule
		
	\end{tabular}
	\caption{Ablation study of different loss functions at TERC 1\% and Irony 1\%. The results are reported by Macro-F1 under five times repeated experiments.}
	\label{tab:loss_fn}
\end{table}
The experimental results of different loss functions at \textbf{TERC} 1\% and \textbf{Irony} 1\% is presented in Tab.~\ref{tab:loss_fn}. The combined loss function Eq.~(\ref{eq-EPiDA-tot}) outperforms Eq.~(\ref{eq-org}) and Eq.~(\ref{eq-aug}) in \textbf{TERC} 1\% and \textbf{Irony} 1\%. As mentioned earlier (\textit{c.f.} Visualization study in main paper), the samples augmented by EPiDA are more diverse than the original samples, which also causes a deviation. Such deviation limits the classification performance. However, the combined loss function Eq.~(\ref{eq-EPiDA-tot}) solved this problem by mixing the augmented samples and the original samples.  
\section{Generation Speed}
 \begin{table}
 \resizebox{0.45\textwidth}{!}{
 	\centering
 \begin{tabular}{cc|cc|c}
 	\toprule
 	EDA & +EPiDA  & CWE & + EiPDA & DataBoost\cr
 	\midrule
 	188.4 & 43.8 &30.7 &10.1 &1.0   \\
 	\bottomrule
 \end{tabular}
 }
 \caption{Generation speed comparison with existing DA methods. The speed is measured by the samples generated per Second. Except for DataBoost whose data are cited from \cite{liu2020data}, all the other methods' results are obtained on a NVIDIA RTX 3090.}
 \label{tab:speed}
 \end{table}
 Tab.~\ref{tab:speed} presents the results of generation speed of EPiDA. We evaluate the speed by the number of samples generated by a DA algorithm per second. As shown in Tab.~\ref{tab:speed}, after using EPiDA, EDA and CWE are still faster than DataBoost.
\section{Effect of the amplification factor $K$.}
 \begin{table}
 \resizebox{0.45\textwidth}{!}{
 	\centering
 	\begin{tabular}{cccccc}
 		\toprule
 		$K$ & 2 & 3  & 5 & 7 & 10\cr
 		\midrule
 		Macro-F1  & 0.573 & 0.577  & 0.576 & 0.574 & 0.575\\
 		\bottomrule
 	\end{tabular}
 	}
 	\caption{Comparison of the classification performance on Irony 1\%
 		under different amplification factor $K$ values.}
 	\label{tab:ab:k}
 \end{table}
By grid search, we present the performance results of different $K$ values in Tab.~\ref{tab:ab:k}, from which we set $K$ to 3 in our experiments.
\section{More Verification of EPiDA}

In order to fully demonstrate the performance of EPiDA, we additionally follow the experimental settings of \cite{shi2021substructure} and compare our method with SUB$^2$. The dataset and classifier in this experiment is SST~\cite{socher2013recursive} and XLM-R~\cite{conneau2019unsupervised}, respectively. Following \cite{shi2021substructure}, to avoid over-fitting to the small development set and tuning on test set issues, we introduce small "development test" (devtest) sets for SST, and only evaluate on the test sets using classifiers with
the best devtest performance. The experimental results are placed in Tab.~\ref{tab:txt-class}.
\begin{table}[t]
	\centering\small
	\begin{tabular}{lr}
		\toprule
		Method & Accuracy\\
		\midrule
		\multicolumn{2}{l}{SST-10\% ~~($|\mathcal{D}_\textit{train}|=0.8\text{K}, |\mathcal{D}_\textit{devtest}|=0.1\text{K}$)}\\
		\midrule 
		NOAUG & 25.4\\
		EDA~\cite{wei2019eda} & 40.6\\
		CWE~\cite{kobayashi2018contextual} & 44.9 \\ 
		SUB$^2$~\cite{shi2021substructure} & 45.8 \\
		EPiDA+EDA & 43.5 \\
		EPiDA+CWE & \textbf{45.9} \\
		
		\bottomrule
	\end{tabular}
	\caption{Accuracy ($\times 100$) on the SST standard test set. The best numbers in each section are bolded. } 
	\label{tab:txt-class}
\end{table}
As shown in Tab.~\ref{tab:txt-class}, after introducing EPiDA, the performance of EDA and CWE are improved. Besides, our method can also achieve comparable performance with SUB$^2$ in SST task, which demonstrates the superiority of our framework.
\begin{table}[h]
	\centering
	\scalebox{1.0}{
		\begin{tabular}{c|cc}
			\hline
			\multirow{2}*{Method} & \multicolumn{2}{c}{Corpus}\\
			\cline{2-3}
			~&AGNews & MR   \\
			\hline
			BERT &0.944 & 0.868 \\
			VDA & 0.945  & 0.878\\
			EPiDA with EDA & \textbf{0.949} & \textbf{0.879}\\
			\hline
	\end{tabular}}
	\caption{Accuracy ($\times 100$) on the test sets of AGNews and MR. The best numbers in each section are bolded.}
	\label{tab:vda}
\end{table}

We also provide the experimental results following the setting of VDA~\cite{zhou2021virtual} in AGNews~\cite{zhang2015character} and MR~\cite{pang2005seeing} corpus. We take BERT as classifier, the experimental results are placed in Tab.~\ref{tab:vda}. As shown in Tab.~\ref{tab:vda}, EPiDA outperforms VDA in classification accuracy.
\section{Apply EPiDA performs in high-resource settings}
\pgfplotsset{width=6.5cm,compat=1.9}
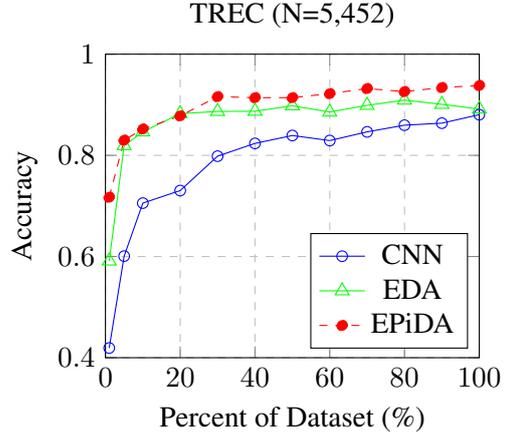
\begin{figure}[!t]
	\begin{centering}
		\begin{tikzpicture}
			\begin{axis}[
				title={TREC (N=5,452)},
				xlabel={Percent of Dataset (\%)},
				ylabel={Accuracy},
				xmin=0, xmax=100,
				ymin=0.4, ymax=1.0,
				xtick={0,20,40,60,80,100},
				ytick={0.4, 0.6, 0.8, 1.0},
				legend pos=south east,
				ymajorgrids=true,
				xmajorgrids=true,
				grid style=dashed,
				]
				\addplot[
				color=blue,
				mark=o,
				mark size=2pt,
				]
				coordinates {
					(1,	    0.4192)
					(5,	    0.6008)
					(10,	0.7056)
					(20,	0.7304)
					(30,	0.7984)
					(40,	0.8236)
					(50,	0.8392)
					(60,	0.8292)
					(70,	0.8464)
					(80,	0.8596)
					(90,	0.8636)
					(100,	0.8804)
				};
				\addlegendentry{CNN}
				\addplot[
				color=green,
				mark=triangle,
				mark size=3pt,
				]
				coordinates {
					(1,	    0.5912)
					(5,	    0.8192)
					(10,	0.8464)
					(20,	0.8824)
					(30,	0.8868)
					(40,	0.8872)
					(50,	0.898)
					(60,	0.8856)
					(70,	0.8988)
					(80,	0.9092)
					(90,	0.9008)
					(100,	0.8916)
				};
				\addlegendentry{EDA}
				\addplot[
				color=red,
				mark=*,
				mark size=2pt,
				dashed,
				]
				coordinates {
					(1,	    0.7172)
					(5,	    0.830)
					(10,	0.852)
					(20,	0.878)
					(30,	0.916)
					(40,	0.914)
					(50,	0.914)
					(60,	0.922)
					(70,	0.932)
					(80,	0.926)
					(90,	0.934)
					(100,	0.938)
				};
				\addlegendentry{EPiDA}
				coordinates {
					(0,	    0.95)
					(100,	0.95)
				};
			\end{axis}
		\end{tikzpicture}
		\caption{Classification accuracy w/o EPiDA for various original data sizes (before DA) used for training.}
		\label{fig:saturation}
	\end{centering}
\end{figure}

In Fig.~\ref{fig:saturation} we provide classification performance vs. original training data size. EPiDA performs well in low-resource settings. However, ever when all the data are used, EPiDA still boosts accuracy (CNN: 0.88, EDA: 0.89, ours: 0.93).
\section{Visualization Case of REM and CEM}
Here we provide the visualization results of using REM and CEM separately to illustrate the benefits of REM and CEM more clearly.
\begin{figure}
	\begin{center}
		\includegraphics[width=\linewidth]{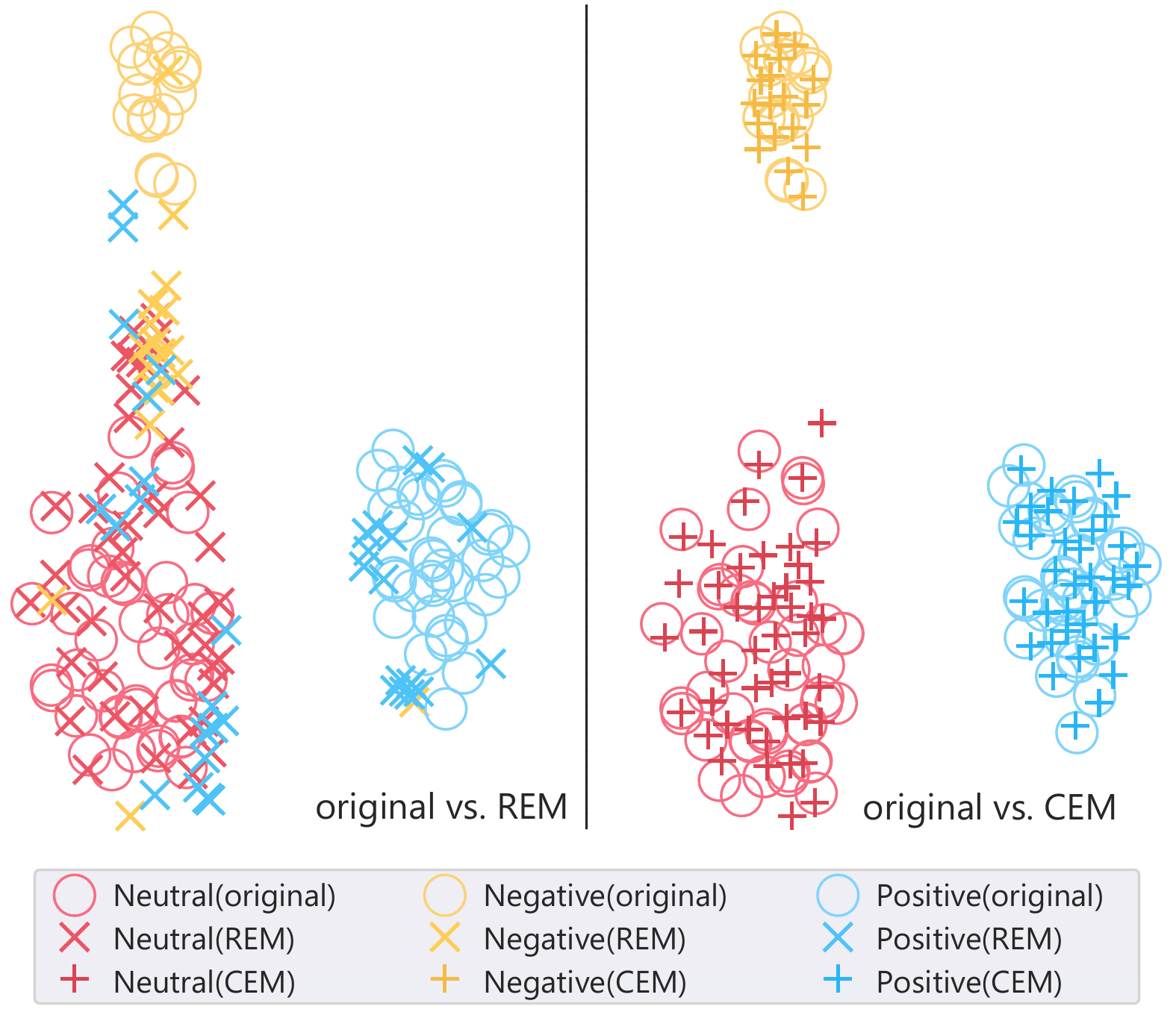}
	\end{center}
	\caption{t-SNE hidden state visualization results of Sentiment Analysis task. Left: original vs. REM only; Right: original vs. CEM only. Different colors represent different classes (Neutral, Negative, and Positive), while different shapes represent different augmentation algorithms (original, REM only and CEM only).}
	\label{fig:supp:gm_mim}
\end{figure}
As shown in Fig.~\ref{fig:supp:gm_mim}, REM encourages to enhance diversity while low-quality samples with wrong labels will be generated. In contrast, CEM encourages to generate high-quality while less-diversity samples. 
\section{Visualization Case of CWE}
\begin{figure}
	\begin{center}
		\includegraphics[width=\linewidth]{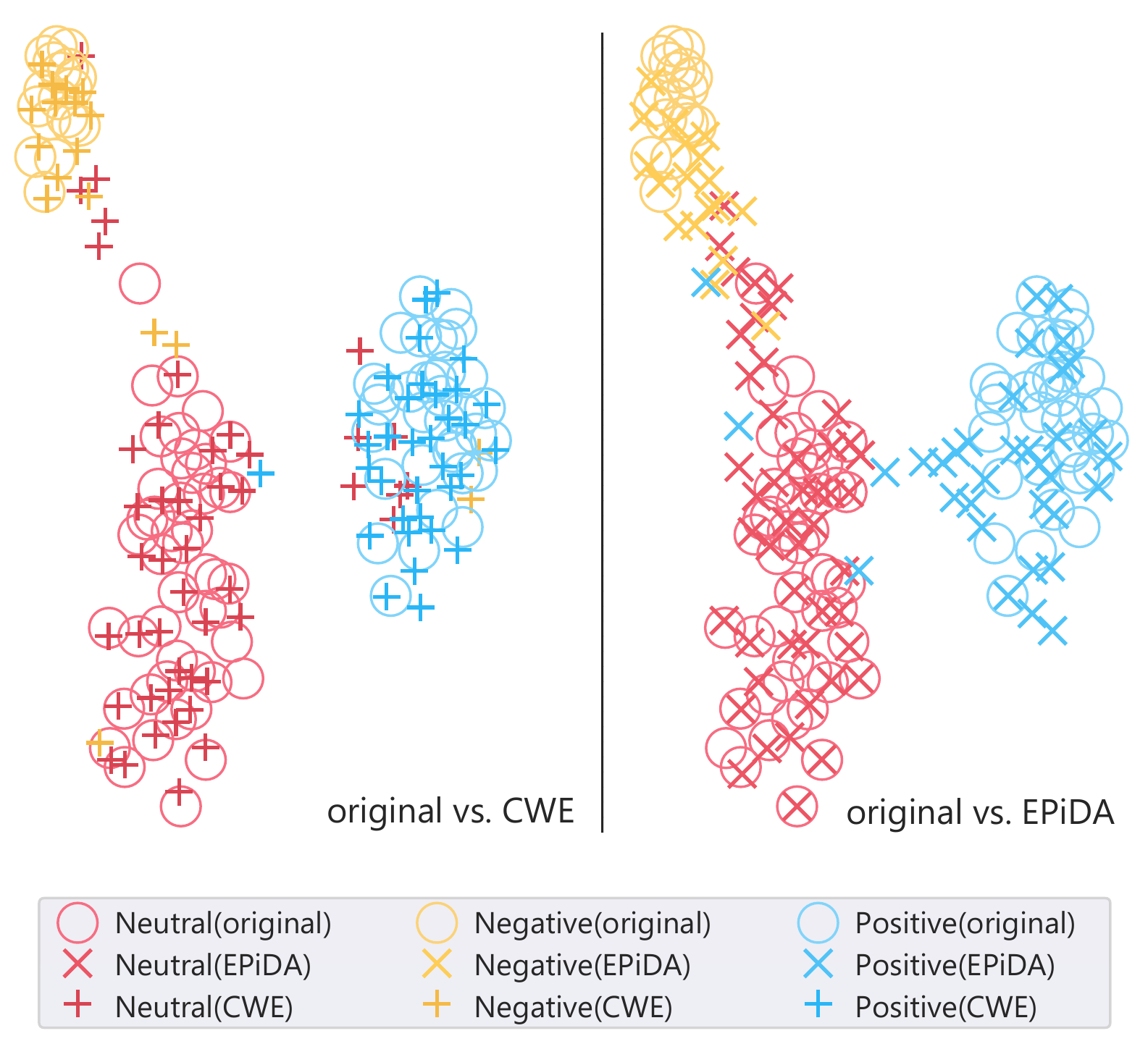}
	\end{center}
	\caption{t-SNE hidden state visualization results of Sentiment Analysis task. Left: original vs. CWE; Right: original vs. EPiDA+CWE. Different colors represent different classes (Neutral, Negative, and Positive), while different shapes represent different augmentation algorithms (original, CWE only and EPiDA+CWE).}
	\label{fig:supp:cwe}
\end{figure}

Fig.~\ref{fig:supp:cwe} show the visualization result of CWE~\cite{kobayashi2018contextual} and EiPDA+CWE. Similar conclusions can also be drawn from Fig.~\ref{fig:supp:cwe}. CWE itself has the ability of enhancing diversity, and with the help of EPiDA, the quality of the generation has been dramatically improved (See Positive Class).
\section{Core Implementation Code}
\begin{figure}[t]
	
	\begin{lstlisting}[style=mystyle]
		EPS = 1e-10
		def REM(z, zt):
		z[(z < EPS).data] = EPS
		return -torch.sum(zt*torch.log(z))
		def MI(z, zt):
		C = zt.size()[1]
		P = (z.unsqueeze(2) * zt.unsqueeze(1)).sum(dim=0)
		P = ((P + P.t()) / 2) / P.sum()
		P[(P < EPS).data] = EPS
		Pi = P.sum(dim=1).view(C, 1).expand(C, C)
		Pj = P.sum(dim=0).view(1, C).expand(C, C)
		return 1.0 - (P * (log(Pi) + log(Pj) - log(P))).sum()
		def H(z):
		z[(z < EPS).data] = EPS
		return -(z*torch.log(z)).sum()
		def CEM(z, zt):
		return MI(z, zt) - H(z)
	\end{lstlisting}
	
	\caption{\label{f:code} Python implementation of REM and CEM. $z$ is the probability distribution predicted by the classifier, and $z_t$ is the probability distribution of original sample.}
\end{figure}
The implementation of REM and CEM is available at Fig.~\ref{f:code}. Here, the calculation of mutual information refers to \cite{ji2019invariant}. 
\section{Replacement of REM and CEM}
Here we discuss the replacement of REM and CEM. In other words, we separately use PPL or cosine similarity mentioned in \cite{zuo2021learnda} to replace REM or CEM to control diversity or quality. The experimental results are presented in Tab.~\ref{tab-rep-rem-cem}. As shown in Tab.~\ref{tab-rep-rem-cem}, REM+CEM outperforms other variants, which demonstrates the superiority of our method.
\begin{table*}
	\centering
	\begin{tabular}{cc|cc|cc}
		\toprule
		REM &PPL&CEM &CosSim & TREC 1\% & Irony 1\% \cr
		\midrule
		$\checkmark$ & - & $\checkmark$ & - & \textbf{0.740} & \textbf{0.576}  \\ 
		-& $\checkmark$ & $\checkmark$ & - & 0.731 & 0.567 \\ 
		$\checkmark$ & - & - & $\checkmark$ & 0.736 & 0.566 \\ 
		- & $\checkmark$ & - & $\checkmark$  & 0.730& 0.562\\ 
		\bottomrule
		
	\end{tabular}
	\caption{Ablation study of the replacement of REM and CEM at TERC 1\% and Irony 1\%. The results are reported by Macro-F1 under five times repeated experiments.}
	\label{tab-rep-rem-cem}
\end{table*}
\section{More Implementation Details}
Here we supply additional details of our implementation.

\textbf{Dataset Preprocessing:} We clean all punctuation, stop words, hashtags, numbers
and URL links in the tweets corpora.

\textbf{Data Augmentation Algorithms:} There are three DA algorithms used in this paper: EDA~\cite{wei2019eda}\footnote{https://github.com/jasonwei20/eda\_nlp}, CWE~\cite{kobayashi2018contextual}\footnote{https://github.com/makcedward/nlpaug}, and TextAttack~\cite{morris2020textattack}\footnote{https://github.com/QData/TextAttack}.

\textbf{Classifiers:} Here we provide the implementation of the classifiers. There are four classifiers used in our paper: CNN~\cite{kim2014convolutional}\footnote{https://github.com/galsang/CNN-sentence-classification-pytorch}, BERT~\cite{devlin2019bert}\footnote{https://huggingface.co/transformers/model\_doc/bert.html}, XLNet~\cite{yang2019xlnet}\footnote{https://huggingface.co/transformers/model\_doc/xlnet.html} and XLM-R~\cite{conneau2019unsupervised}\footnote{https://huggingface.co/transformers/model\_doc/xlmroberta.html}.

\textbf{Random Seeds:} The random seeds used in this paper are 0,1,2,3 and 4, respectively.

\textbf{Others:} We take AdamW~\cite{loshchilov2018fixing} as the optimizer. All the experiments are conducted at 4 NVIDIA RTX 3090 GPUs with Pytorch1.8.
\section{Limitation}
The major limitation of EPiDA is the training time. Although EPDA can bring performance improvements, it will reduce the training speed by at least $K$(the amplification factor) times. This means that when the DA method and the classifier itself are cumbersome, the overall training time will be long.
Besides, how to measure or define samples' value is still an open problem. 

\section{Supplementary Example}
In Tab.~\ref{tab-example}, we provide several detailed augmentation results of EPiDA. $m$ and $K$ are set to 3. Therefore, 9 candidate samples will be generated.
\begin{table*}
	\centering
	\begin{tabular}{c@{\hspace{0.4em}}|c|c@{\hspace{0.4em}}c@{\hspace{0.4em}}c@{\hspace{0.4em}}}
		\toprule[2pt]
		Task/Selected &Sentence & $s_{div}$ & $s_{qua}$ & $s_{tot}$  \cr
		\midrule[1pt]
		Sentiment & \tabincell{c}{I'm about to eat four hot dogs and watch Miss USA. \underline{Happy} Sunday.}& 0.00 & 1.00 &1.00 \\
		\midrule
		&\tabincell{c}{I am about to eat four hot dogs and watch Miss usa. \underline{Happy} Sunday.}& 0.10 & 0.83 &0.93 \\
		&\tabincell{c}{I'm about to eat four track hot dogs and watch Miss USA. \underline{Happy} Sunday.}& 0.17 & 0.70 &0.87 \\
		$\checkmark$&\tabincell{c}{I'm about to eat four hot dogs and watch Miss USA. \underline{Happy} Sunday.}& 0.00 & 1.00 &1.00 \\
		$\checkmark$&\tabincell{c}{I'm about to eat four hot live dogs and watch Miss USA. \underline{Happy} Sunday.}& 0.06 & 0.98 &1.04 \\
		&\tabincell{c}{I'm about to eat four hot dogs and watch Miss USA. Gold Sunday.}& 0.24 & 0.58 &0.82 \\
		&\tabincell{c}{I'm about to eat four hot dogs and watch Sun Miss USA. \underline{Happy} Sunday.}& 0.00 & 0.98 &0.98 \\
		$\checkmark$&\tabincell{c}{I'm about to eat hot dogs and watch Miss USA. \underline{Happy} Sunday.}& 0.03 & 0.99 &1.02 \\
		&\tabincell{c}{I'm about to eat quadruplet hot dogs and watch Miss USA. Sunday.}& 1.00 & 0.00 &1.00 \\
		&\tabincell{c}{I'm about to eat four hot dogs and watch Miss USA. \underline{Happy} Sunday.}& 0.00 & 1.00 &1.00 \\
		\midrule[2pt]
		Irony & \tabincell{c}{A wonderful day of starting \underline{work} at 6am }& 0.00 & 1.00 &1.00 \\
		\midrule
		& \tabincell{c}{A day wonderful of starting \underline{work} at 6am}& 0.43 & 0.48 &0.91 \\
		& \tabincell{c}{A 6am day of starting \underline{work} at wonderful}& 0.21 & 0.73 &0.94 \\
		& \tabincell{c}{A wonderful of starting \underline{work} at 6am}& 0.75 & 0.18 &0.93 \\
		& \tabincell{c}{A grand day of starting \underline{work} at 6am}& 0.39 & 0.53 &0.92 \\
		$\checkmark$& \tabincell{c}{Day wonderful a of starting \underline{work} at 6am}& 0.92 & 0.07 &0.99 \\
		& \tabincell{c}{A wonderful day starting of \underline{work} at 6am}& 0.56 & 0.35 &0.91 \\
		$\checkmark$& \tabincell{c}{A wonderful day of starting at 6am}& 1.00 & 0.00 &1.00 \\
		$\checkmark$& \tabincell{c}{A day of starting \underline{work} at 6am}& 0.87 & 0.12 &0.99 \\
		& \tabincell{c}{A wonderful day at starting \underline{work} of 6am}& 0.75 & 0.18 &0.93 \\
		\bottomrule[2pt]
	\end{tabular}
	\caption{Some examples selected by SEAS. Underlined words are salient words. The first column will be checked if this augmented sample is seleceted by SEAS.}
	\label{tab-example}
\end{table*}


\end{document}